\documentclass{article}

\usepackage{amsfonts}
\usepackage{amsmath}
\usepackage{caption}
\usepackage{graphicx}
\usepackage[utf8]{inputenc}
\usepackage[preprint,nonatbib]{neurips_2021}
\usepackage{subcaption}
\usepackage{url}

\newcommand{\E}{\mathbb{E}}

\newcommand{\R}{\mathbb{R}}

\newcommand{\capwidth}{1}

\newcommand{\omitt}[1]{}

\title{Rotating spiders and reflecting dogs: a class conditional approach to learning data augmentation distributions}
\author{
  Scott Mahan \\ 
  Department of Mathematics \\
  University of California, San Diego \\
  La Jolla, CA 92093 \\
  \texttt{scmahan@ucsd.edu} \\
  \And
  Henry Kvinge \\
  Pacific Northwest National Laboratory \\
  Seattle Research Center \\
  Seattle, WA 98109 \\
  \texttt{henry.kvinge@pnnl.gov}
  \And
  Tim Doster \\
  Pacific Northwest National Laboratory \\
  Seattle Research Center \\
  Seattle, WA 98109 \\
  \texttt{timothy.doster@pnnl.gov}
}
\date{}

\begin{document}

\maketitle

\begin{abstract}
    Building invariance to non-meaningful transformations is essential to building efficient and generalizable machine learning models. In practice, the most common way to learn invariance is through data augmentation. There has been recent interest in the development of methods that learn distributions on augmentation transformations from the training data itself. While such approaches are beneficial since they are responsive to the data, they ignore the fact that in many situations the range of transformations to which a model needs to be invariant changes depending on the particular class input belongs to. For example, if a model needs to be able to predict whether an image contains a starfish or a dog, we may want to apply random rotations to starfish images during training (since these do not have a preferred orientation), but we would not want to do this to images of dogs.  In this work we introduce a method by which we can learn class conditional distributions on augmentation transformations. We give a number of examples where our methods learn different non-meaningful transformations depending on class 
    and further show how our method can be used as a tool to probe the symmetries intrinsic to a potentially complex dataset. 
\end{abstract}

\section{Introduction}

Most real world data contains an enormous amount of information that is not related to the task of interest. Because of this, a model's ability to be invariant to a range of non-meaningful variation in input is essential to its robustness. This is particularly critical for models that will be deployed to the real world, where we need to be confident that a model will not fail just because input has been superficially transformed \cite{hendrycks2020many,hendrycks2019benchmarking}.

The conventional method of building invariance into a model is through data augmentation, where one synthetically alters data at training time to simulate the range of variation that the model is expected to see at inference time. While augmentation strategies are generally developed in an {\emph{ad hoc}} manner, using a few standard augmentations associated with a data modality, there has recently been effort to make data augmentation more responsive to the model and data itself \cite{CZMVL19,LKKKK19,vBJH18}. One recent method called Augerino \cite{BFIW20} learns a distribution of augmentations that actually reflects the extent to which those augmentations appear in data. In principle, if dogs and cats appear at any orientation in an image classification dataset, then when Augerino is applied to rotation transformations, it will allow all possible rotations to be applied during training. On the other hand, if dogs and cats only appear at angles from $-90^\circ$ to $90^\circ$ (that is, dogs and cats lay on their sides but never upside down), then Augerino will learn to only apply this range of random rotations during training. 

Of course, there are many real-world situations where the augmentations that should be applied to input depend heavily on the class to which that input belongs. As suggested by our title, if one is trying to differentiate images containing spiders from those containing dogs, one would probably want to apply a different range of augmentations to each. Spiders can naturally appear at many orientations so it would make sense to apply random rotations to them. Dogs on the other hand are generally not seen upside down, so it would make less sense to waste the model's capacity learning to identify them under this transformation. To our knowledge, this is the first time that the need for class conditional augmentation has been identified in the literature.

To mitigate in this issue, in this paper we propose a class conditional method of learning how to augment data during training. Our method builds off the success of Augerino, expanding the single learned augmentation distribution used there to a full set of learned augmentation distributions, one for each class. We show that our method is more robust than Augerino to invariances in the data 
in a range of settings, both synthetic and naturally occuring, using the MNIST \cite{LCB10}, CIFAR-10, and CIFAR-100 datasets \cite{K09}. Taken together, the work presented here represents another step in the process of making data augmentation actually reflect the nature of the data with which we want to train our models.

\subsection{Related Work}

There are two approaches to building invariances into machine learning models: hardcoding invariances in network architectures and learning invariances through training. As mentioned in the Introduction, the primary example of the former are convolutional neural networks (CNNs) which achieve translation invariance through the combination of translation equivariant convolutional layers and pooling \cite{LB98}. Building on this success, there has recently been a surge of interest in neural networks that incorporate invariance or equivariance to other types of transformations in their architectures. This can be done by pooling outputs over fixed transformations of inputs \cite{LSBP16}, applying transformed convolutional filters to input data \cite{MVKT17, ZYQJ17}, or using steerable filters \cite{CW16a, CW16b, SSS20, WC19, WW19, WGTB17}. All of these approaches assume prior knowledge of invariance to a full augmentation distribution over the data (e.g., invariance to rotation by any angle).

In practice, a more common approach is learning invariance by augmenting training data with transformed images. While the standard approach is to pick a few basic transformations to apply to input data at training time, there have been significant efforts to (1) identify the specific transformations that will most improve a model's performance or (2) the distribution representing the types of variation that are actually present in the data. AutoAugment \cite{CZMVL19} uses reinforcement learning to search for an optimal augmentation policy over a discrete space, hence automatically discovering invariance in the data. Fast AutoAugment \cite{LKKKK19} reduces the extreme computation time of AutoAugment by using a density matching algorithm. Both of these algorithms leverage reinforcement learning which uses validation loss as a reward for learning the augmentation distribution. Faster AutoAugment \cite{HZYN19} and Differentiable Automatic Data Augmentation \cite{LHWHRY20} learn augmentation distributions more efficiently using backpropagation, but still rely on validation loss.

While the methods above attempt to learn optimal augmentation using a strategy based on model performance, another approach is to simply try to learn the distribution of augmentations present in the training data itself. One approach involves learning parameterized invariances using the marginal likelihood of a Gaussian process \cite{vBJH18}. The method Augerino \cite{BFIW20}, which was mentioned in the Introduction, uses regularization to encourage data augmentation without sacrificing model accuracy, thus learning the largest distribution of non-meaningful augmentations from training data alone.

None of these previous methods take into account that for many tasks, the distribution of augmentation transformations may differ radically for different classes. For example, certain class labels may exhibit invariances (e.g., rotation invariance of overhead images) while other classes may not. We focus on this setting and extend the Augerino framework to allow for class conditional invariances that further enhance model generalization and robustness.

\subsection{Contributions}

Our main contribution is extending the Augerino framework to the setting of class conditional invariance. To our knowledge, the problem of learning class conditional invariance from training data has not been studied despite the fact that it appears naturally in many datasets. After establishing a mathematical framework for a class conditional version of Augerino, we conduct experiments on modified versions of MNIST, CIFAR-10, and CIFAR-100. Our class conditional model is able to accurately learn augmentation distributions over non-meaningful transformations for different classes, and as a result helps to train a model that ignores certain kinds of variation in the settings that they actually appear.

\omitt{
\subsection{Outline of the Paper}

Our work discusses the mathematical background for learning class conditional invariances from training data, then presents experimental results showing the effectiveness of our class conditional model. Section \ref{sec:Augerino} outlines the Augerino framework for learning invariances from training data and extends the model to the setting of class conditional invariance in Section \ref{subsec:AugerinoClass}. Section \ref{sec:experiments} discusses our experiments with learning class conditional invariance, including datasets synthetically designed to have class conditional invariance in Sections \ref{subsec:RotMNIST} and \ref{subsec:ColCIFAR} and a dataset with natural class conditional invariance in Section \ref{subsec:RotCIFAR}. Section \ref{sec:limitations} discusses some limitations of our work and possible ways to address them, and Section \ref{sec:conclusions} draw conclusions from our results. 
}

\section{Learning Invariance} \label{sec:Augerino}

To learn invariance from training data, the Augerino model introduced in \cite{BFIW20} samples data augmentations from a uniform distribution over some parameterized space of transformations and feeds the transformed input through a multilayer perception or a convolutional neural network. During training, each input is transformed once and the augmentation parameter is learned along with the network weights. For testing data, each input is augmented with several transformed copies that are all fed through the network. The outputs of all transformed inputs are averaged to obtain a final result. Thus, the Augerino model is
\[
    f(x;\theta,w) = \E_{g \sim \mu_\theta} \varphi(gx;w)
\]
where $\varphi(\cdot;w)$ is a neural network with weights $w$ and $\mu_\theta$ is a uniform distribution over some set of transformations parameterized by $\theta$ (e.g., rotation of an image by some angle in $[-\theta,\theta]$). The expectation over $\mu_\theta$ is approximated by averaging the outputs of a finite number of transformed inputs. 

To train the augmentation and network parameters in \cite{BFIW20}, the authors use cross-entropy loss on the log probabilities given by the model coupled with a regularization term that encourages augmentation. Specifically, the loss function is
\[
    L(\theta,w;x) = \E_{g \sim \mu_\theta} \ell(\varphi(gx;w),y) - \lambda \|\theta\|^2
\]
where $\ell(\varphi(gx;w),y)$ is the cross-entropy loss given the output log probabilities $\varphi(gx;w)$ and the true class label $y$ of input $x$ and $\lambda$ is an augmentation regularization parameter. When training the model in \cite{BFIW20}, $g \sim \mu_\theta$ is sampled using a reparameterization trick from \cite{KW13} as $g=\theta\epsilon$ with $\epsilon$ sampled from a uniform distribution $U[-1,1]$. This allows for $\theta$ to be treated as a learnable parameter. 

Typically, learning invariance from training data is challenging because widening the augmentation distribution enhances generalization but does not decrease training loss. The regularization term encourages augmentation up until the point where the cross-entropy loss increases due to too much augmentation. Training this model typically involves applying transformations to data that have already been augmented. This helps the model generalize to invariance in the data and learn when augmentation becomes harmful to accuracy. During testing, we apply transformations to several copies of inputs that are already augmented according to the natural invariance already present in the test set. This may result in feeding the network transformed inputs that lie outside the desired set of invariant transformations. However, since we average over several copies and it is likely that most transformed copies lie in the support of the augmentation distribution $\mu_\theta$, the model can augment inputs without negatively affecting accuracy.

Augerino can incorporate $d$ types of transformations (e.g., rotation, scaling, and shearing) by simply allowing the augmentation parameter $\theta$ to lie in $\R^d$. The parameter corresponding to each transformation can be trained separately to allow for very rich data augmentation. However, one limitation of Augerino is that each input is transformed according to the same augmentation distribution. In the real world however, it is easy to find problems where augmentations that make sense for one class would not make sense for another. In Section \ref{subsec:RotCIFAR} we provide of an example of this using CIFAR-100.

\subsection{Class Conditional Invariance} \label{subsec:AugerinoClass}

Ignoring class when considering invariance can result in either learning insufficient invariance for some classes or learning too much invariance for others. To model class conditional invariance, we propose a version of Augerino with a full set of $d$ augmentation parameters for each of $K$ classes. Thus, the model becomes
\[
    f(x,k;\Theta,w) = \E_{g \sim \mu_{\theta_k}} \varphi(gx;w)
\]
where $\Theta \in \R^{d \times K}$ is the full set of augmentation parameters and $\theta_k \in \R^d$ parameterizes the augmentation distribution for just class $k$. In practice, we compute $\theta_k = \Theta e_k$ where $e_k \in \R^K$ is a one-hot encoded class label corresponding to class $k$. Conditioning on class label in this way still allows for backpropagation to update the augmentation parameters. Overall, this model trains a single network $\varphi$ on the augmented inputs, while efficiently conditioning the augmentation distribution on the input's class label. For input $x$ belonging to class $k$, the model's loss function becomes
\[
    L(\Theta,w;x) = \E_{g \sim \mu_{\theta_k}} \ell(\varphi(gx;w),y) - \frac{\lambda}{K} \sum_{k=1}^K \|\theta_k\|^2.
\]
Dividing by the number of classes $K$ ensures that the amount of regularization in the model is similar to that of the standard Augerino model with the same regularization parameter $\lambda$.

\paragraph{Class Conditional Augmentation During Testing} When training our model, class labels are given and can be used to determine the augmentation distribution for each input. However, during testing, class labels are unknown and the model does not know how to transform the input. To address this challenge, we feed the non-transformed input through the network $\varphi$ and obtain a predicted class label from the output $\varphi(x;w)$. Since the network is trained on augmented data, it should give reliable predictions even when the test input $x$ is already transformed according to some symmetry naturally present in the dataset. For example, if the model receives an image of a dog already rotated at $45^\circ$, then rotating it again between $-90^\circ$ and $90^\circ$ may result in the dog appearing upside down. However, Augerino and our class conditional model transform several copies of the input, and on average we expect to see at least half of the transformations in a natural rotation range. 

\section{Experiments} \label{sec:experiments}

To illustrate the applicability and success of neural networks with class conditional invariance, we conduct several experiments training our extension of Augerino on data with different levels of invariance depending on class label. In each experiment, the training data exhibit invariance that is either synthetically or naturally class-dependent. We compare our class conditional Augerino model to a standard Augerino model, showing that our model is able to successfully learn different augmentation parameters for each class in a way that enhances robustness 
on the testing data. We implemented our class conditional models by adding Python classes to the Augerino code repository created by Benton et al. \cite{BFIW20}, available at \url{https://github.com/g-benton/learning-invariances}. Models were trained using about 100 hours on NVIDIA V100 GPUs with XSEDE resources \cite{XSEDE}. 

\subsection{Class Conditional Rotation Invariance: Rotating MNIST by Class} \label{subsec:RotMNIST}

As a motivating example, we first consider a rotated version of the MNIST dataset \cite{LCB10}, where each image is rotated by a random angle in a range conditional on its class label. In our first experiment, all digits are randomly rotated in the full range of $[-\pi,\pi]$ except for 6's and 9's, which are only rotated in the range $[-\pi/4, \pi/4]$. We believe this is a natural setup for a class conditional invariance problem, since 6's and 9's can be confused for each other when rotated too much but all other digits are distinct, regardless of their orientation. Hence, we want a model that can be fully invariant to rotation on most classes to improve generalization, but that limits invariance on 6's and 9's so as not to harm prediction accuracy. 

We train a standard Augerino model and our class conditional Augerino model on this dataset for 10 training epochs each using an Adam optimizer \cite{KB17} with a learning rate of $10^{-3}$ and weight decay of $10^{-4}$. Both models use an augmentation regularization parameter of $0.05$, a 5-layer convolutional neural network with one fully connected layer, and 4 transformed copies of each input during testing.

\begin{figure}[ht]
	\centering
	\begin{subfigure}[b]{0.495\linewidth}
	    \centering
	    \includegraphics[width=1.0\linewidth]{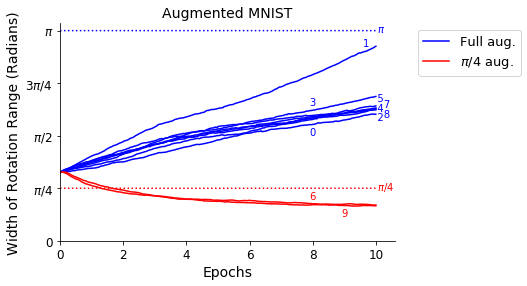}
	    \caption{Augmentation parameters. \label{subfig:RotMNISTaug}}
	\end{subfigure}
	\begin{subfigure}[b]{0.495\linewidth}
	    \centering
	    \includegraphics[width=1.0\linewidth]{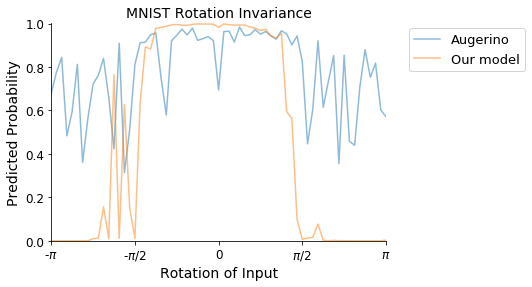}
	    \caption{Model invariance. \label{subfig:RotMNISTinv}}
	\end{subfigure}

	\captionsetup{width=\capwidth\linewidth}
	\caption{We train standard Augerino and our class conditional model on rotated MNIST data with 6's and 9's partially augmented and all other digits fully augmented. (\subref{subfig:RotMNISTaug}) Learned augmentation parameters over 10 training epochs. Solid lines show learned parameters and dashed lines show the width of the distribution used to augment training data for classes of the corresponding color. Blue lines correspond to classes that were rotated in the range $[-\pi,\pi]$. Red lines correspond to 6's and 9's that were rotated in the range $[-\pi/4,\pi/4]$.  (\subref{subfig:RotMNISTinv}) The predicted probability of the correct class label vs.\ orientation for an image of a 6. At orientations in $[-\pi,-\pi/2] \cup [\pi/2,\pi]$, our model gives a high probability that the image is a 9 (not shown).} \label{fig:RotMNIST}
\end{figure}

Figure \ref{fig:RotMNIST} compares the standard and class conditional Augerino models on the conditionally rotated MNIST data. Our class conditional model learns that 6's and 9's are augmented within the range $[-\pi/4,\pi/4]$, while other digits are rotated by larger angles. We see that the width of rotation for 1's grows very quickly, which may be due to 1's being easily identifiable at any orientation.

To illustrate how our model captures the invariance present in the data, we use a test image of a 6 and rotate it at 32 evenly spaced orientations in the range $[-\pi,\pi]$. At each orientation we plot both models' predicted probabilities that the image is a 6. We see that our class conditional model more confidently predicts the image as a 6 within and slightly beyond the augmentation range of $[-\pi/4,\pi/4]$ using during training. At orientations beyond $\pi/2$, our model stops predicting the image as a 6 (although it is not shown in the plot, we note that our model predicts the image as a 9 beyond rotations of $\pi/2$). 

On the other hand, the Augerino model learns to rotate data at angles less than $2.08$ radians, or approximately in the range $[-2\pi/3,2\pi/3]$, well beyond the augmentation range for 6's. As a result, Augerino is less confident than our model when predicting 6's at ``upright'' orientations, and often continues labeling the image as a 6 beyond rotations of $\pi/2$. In this sense, Augerino has learned ``too much'' invariance and may be more likely to mix up 6's and 9's. When we test each model on MNIST testing data that exhibits the same class conditional invariance as the training data, the standard Augerino model has an accuracy of 96.90\% and our class conditional model has an accuracy of 97.49\%. The class conditional model generalizes better to this test set because it can more precisely learn the invariant structure of the data.

We conduct another experiment to investigate our class conditional model handles slight differences in augmentation parameters for each class. In this case, we augment classes by randomly rotating by angles in the following ranges: 1's, 4's, and 7's in $[-\pi/3,\pi/3]$; 2's, 5's, and 8's in $[-\pi/4, \pi/4]$; and 3's, 6's, and 9's in $[-\pi/6,\pi/6]$ (0's are not rotated). By adding these class conditional invariances in the training data, we want to see if our class conditional model recognizes small differences in invariance present in the training data, or rather learns the largest amount of invariance for each class that does not negatively impact model performance. In particular, we do not apply any rotation to 0's since they already exhibit rotation invariance. We again train a standard Augerino and our class conditional model on this data with the same network architecture and training hyperparameters.

\omitt{
\begin{figure}[ht]
	\centering
	\begin{subfigure}[b]{0.495\linewidth}
	    \centering
	    \includegraphics[width=1.0\linewidth]{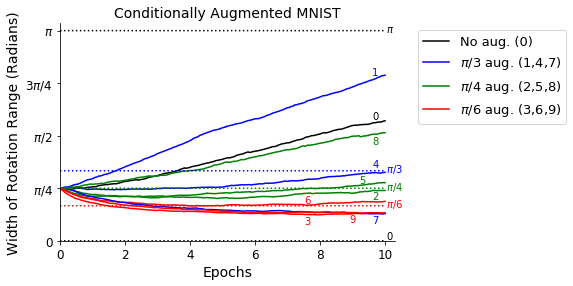}
	    \caption{Augmentation parameters. \label{subfig:RotMNISTaugpar}}
	\end{subfigure}
	\begin{subfigure}[b]{0.495\linewidth}
	    \centering
	    \includegraphics[width=\linewidth]{TempFigures/invariance_rotamt_new.png}
	    \caption{Model invariance. \label{subfig:RotMNISTinvpar}}
	\end{subfigure}

	\captionsetup{width=0.85\linewidth}
	\caption{We compare class conditional models trained on fully augmented MNIST and partially augmented MNIST. (\subref{subfig:RotMNISTaugpar}) Learned augmentation parameters over 10 training epochs. Line colors indicate how much training data was augmented for each class. Dashed lines show the width of the distribution used to augment training data for classes of the corresponding color. (\subref{subfig:RotMNISTinvpar}) Test accuracy on 6's vs. orientation of the MNIST testing data.}\label{fig:RotMNISTpar}
\end{figure}
}

\begin{figure}[ht]
    \centering
    \includegraphics[width=0.75\linewidth]{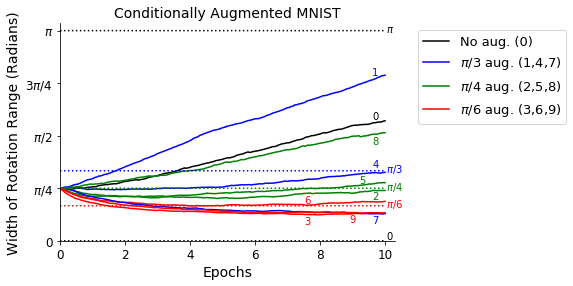}
    
    \captionsetup{width=\capwidth\linewidth}
    \caption{Learned augmentation parameters over 10 training epochs for our class conditional model trained on MNIST data rotated depending on class. Solid lines show learned parameters and dashed lines show the width of the distribution used to augment training data for classes of the corresponding color. The black line corresponds to 0's which were not rotated. Blue lines correspond to classes that were rotated in the range $[-\pi/3,\pi/3]$. Green lines correspond to classes that were rotated in the range $[-\pi/4,\pi/4]$. Red lines correspond to classes that were rotated in the range $[-\pi/6,\pi/6]$.}
    \label{fig:RotMNISTpar}
\end{figure}

Figure \ref{fig:RotMNISTpar} demonstrates how our class conditional model learns different augmentation parameters for each class. For the most part, the model learns smaller augmentation parameters for class labels with less invariance in the training data. For example, it learns small augmentation parameters for 3's, 6's, and 9's which were rotated in the range $[-\pi/6,\pi/6]$ during training, and it learns an augmentation distribution of $[-\pi/4,\pi/4]$ for 2's and 5's. However, this doesn't hold for all classes. In most cases the learned parameter is larger than the amount of invariance used to construct the training data, perhaps due to natural rotations in handwritten digits. We see that the model learns much larger than expected augmentation parameters for 0's, 1's, and 8's, which may be due to these classes exhibiting natural rotation invariance. It is unclear why the model learns such a small augmentation parameter for 7's. Overall, these results seem to indicate that our class conditional model often detects which classes exhibit more or less invariance, but sometimes increases the augmentation parameter when doing so will not impact performance. This makes sense because of the augmentation regularization that encourages increasing the augmentation parameter, but it is also worth noting that the amount of invariance built into the training data affects the learned parameter.

By comparison, the standard Augerino model learns to rotate data at angles less than $0.95$ radians, or approximately in the range $[-3\pi/10,3\pi/10]$. This is a much narrower augmentation distribution than the previous experiment, probably because this model is trained on data with much smaller rotations. In the previous experiment, the Augerino model's accuracy suffered from learning an augmentation distribution that was too wide for 6's and 9's. In this experiment, Augerino does not apply harmful transformations to these classes. Both models generalize well -- on test data that exhibits the same amount of invariance used to train these models on small rotations, Augerino has 98.55\% accuracy and the class conditional model has 98.18\% accuracy. However, our model provides more insight on the invariances present in the data.

\subsection{Class Conditional Brightness Invariance} \label{subsec:ColCIFAR}

We go beyond the setting of rotation invariance by synthetically modifying the brightness of some classes in the CIFAR-10 dataset \cite{K09}. In this experiment, we change the brightness of the airplane, bird, cat, deer, and ship classes by sampling a value uniformly from $[-0.2,0.2]$ for each image and adding this value to each channel and pixel of the image, clamped so that each resulting pixel value is in $[0,1]$. Images from the other classes (automobile, dog, frog, horse, and truck) are not modified. The brightness change is a differentiable transformation that can be learned using the reparameterization trick. We learn the augmentation parameter as $\tilde{\theta} = 0.3f(\theta)$ where $f$ is a sigmoid function to keep the brightness changes in a reasonable range.

\begin{figure}[ht]
	\centering
	\begin{subfigure}[b]{0.32\linewidth}
	    \centering
	    \includegraphics[width=0.9\linewidth]{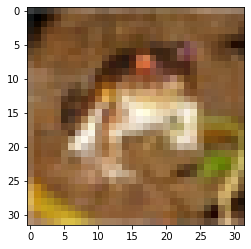}
	    \caption{Original image. \label{subfig:frog}}
	\end{subfigure}
	\begin{subfigure}[b]{0.32\linewidth}
	    \centering
	    \includegraphics[width=0.9\linewidth]{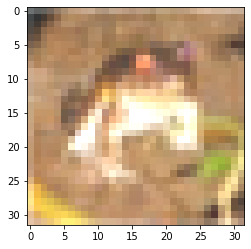}
	    \caption{Increased brightness. \label{subfig:frog_high}}
	\end{subfigure}
	\begin{subfigure}[b]{0.32\linewidth}
	    \centering
	    \includegraphics[width=0.9\linewidth]{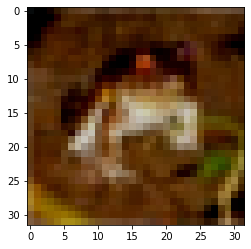}
	    \caption{Decreased brightness. \label{subfig:frog_low}}
	\end{subfigure}

	\captionsetup{width=\capwidth\linewidth}
	\caption{We show brightness augmentations for an example image from CIFAR-10. 
	(\subref{subfig:frog}) The original image. (\subref{subfig:frog_high}) The image with brightness increased by $0.2$ on a $[0,1]$ scale. (\subref{subfig:frog_low}) The image with brightness decreased by $0.2$ on a $[0,1]$ scale.} \label{fig:frog}
\end{figure}

Figure \ref{fig:frog} shows a CIFAR-10 image and the effects of increasing and decreasing its brightness by $0.2$. We train a standard Augerino model and our class conditional model on this dataset with the same model architecture and training hyperparameters as the experiments in Section \ref{subsec:RotMNIST}.

\begin{figure}[ht]
	\centering
	\begin{subfigure}[b]{0.495\linewidth}
	    \centering
	    \includegraphics[width=1.0\linewidth]{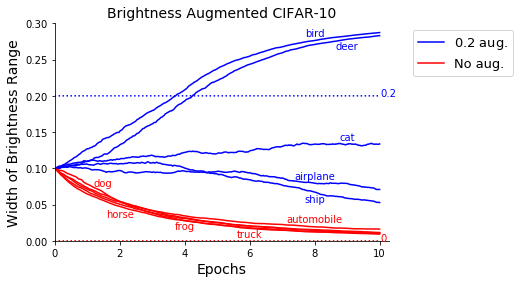}
	    \caption{Augmentation parameters. \label{subfig:ColCIFARaug}}
	\end{subfigure}
	\begin{subfigure}[b]{0.495\linewidth}
	    \centering
	    \includegraphics[width=1.0\linewidth]{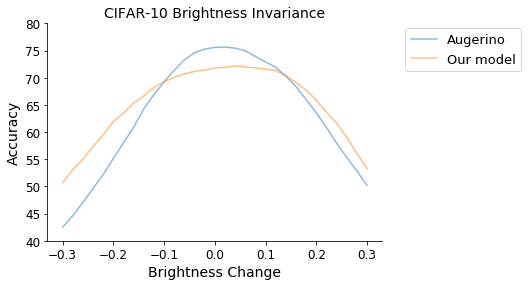}
	    \caption{Model invariance. \label{subfig:ColCIFARinv}}
	\end{subfigure}

	\captionsetup{width=\capwidth\linewidth}
	\caption{We train standard Augerino and our class conditional model on CIFAR-10 images with different brightness invariance by class. (\subref{subfig:ColCIFARaug}) Learned augmentation parameters over 10 training epochs. Blue lines correspond to classes whose brightness was augmented randomly in the range $[-0.2,0.2]$. Red lines correspond to other classes that were not augmented. Dashed lines show the width of the distribution used to augment training data for classes of the corresponding color. (\subref{subfig:ColCIFARinv}) Test accuracy vs.\ brightness changes on the CIFAR-10 testing data.}\label{fig:ColCIFAR} 
\end{figure}

Figure \ref{fig:ColCIFAR} shows the augmentation parameters learned by the models and compares model performance and invariance. We can see that our class conditional model can differentiate between classes that have not had their brightness augmented and classes that have. The model learns small augmentation parameters for classes that were not augmented during training. Interestingly, for classes that were augmented with brightness changes in the range $[-0.2,0.2]$, the model learned larger than expected parameters for the bird and deer classes and smaller than expected parameters for the others. This may occur due to natural brightness invariances present in bird and deer images, or it may be the case that model accuracy on these classes is not harmed by changes in brightness.

To assess each model's brightness invariance, we test them on the entire CIFAR-10 test set with 31 evenly spaced brightness changes between $-0.3$ and $0.3$. We see that Augerino performs better on the unmodified CIFAR-10 data and under small brightness changes, but our class conditional model is more robust to the entire test set changing in brightness. The Augerino model learns to change brightness in the range $[-0.21,0.21]$, so it may be applying too much augmentation to images that already have increased or decreased brightness. The Augerino model is more accurate on a test set that exhibits that same invariances as the training data ($[-0.2,0.2]$ brightness augmentation on some classes and none on others), with 77.27\% compared to 73.39\% for our class conditional model. Our model better reflects the invariance present in the training data, but it may be the case for this dataset that applying extra augmentation to some classes is not harmful.

\subsection{Learning Natural Class Conditional Rotation Invariance} \label{subsec:RotCIFAR}

In the experiments in Sections \ref{subsec:RotMNIST} and \ref{subsec:ColCIFAR}, we synthetically designed class conditional invariance in the training set by augmenting some classes more than others. Now, we want to examine a case where class conditional invariance is already naturally present in the dataset with no augmentation. For this experiment, we hand-pick 10 classes from the CIFAR-100 dataset \cite{K09} and do not modify any of the images. Five of the classes (beetle, clock, crab, flatfish, and sunflower) because we expect them to inherent exhibit rotation invariance since their features do not change much under rotation and in some cases are pictured from overhead. The other five classes (house, plain, road, sea, and skyscraper) were chosen because they are typically pictured at the same orientation in all images, so we expect they do not exhibit any rotation invariance at all.

\begin{figure}[ht]
	\centering
    \includegraphics[width=0.9\linewidth]{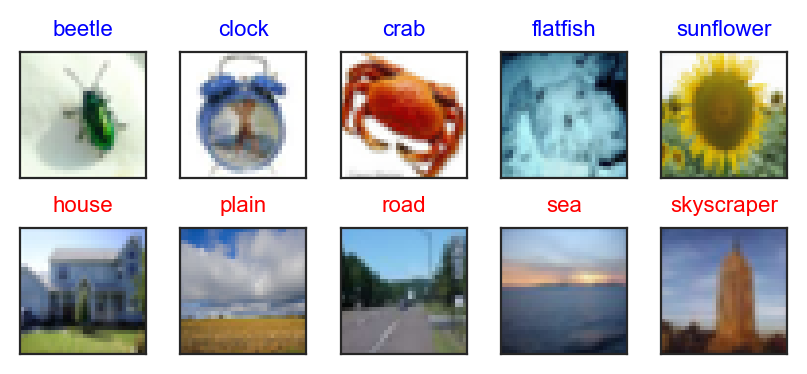}
	    
	\captionsetup{width=\capwidth\linewidth}
	\caption{One image from each selected class of the CIFAR-100 dataset. The classes in the top row were selected because we expect them to naturally exhibit rotational symmetry. The classes in the bottom row are not expected to exhibit any rotation invariance.} \label{fig:CIFAR-100}
\end{figure}

Figure \ref{fig:CIFAR-100} shows some CIFAR-100 images from the selected classes. We train a standard Augerino model and our class conditional model on this dataset. We use the same architecture and training hyperparameters as the experiments in Section \ref{subsec:RotMNIST} but we train for 50 epochs.

\begin{figure}[ht]
	\centering
	\begin{subfigure}[b]{0.495\linewidth}
	    \centering
	    \includegraphics[width=1.0\linewidth]{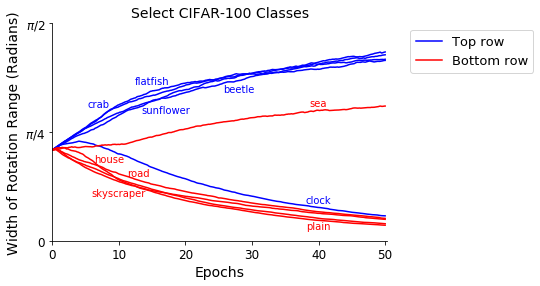}
	    \caption{Augmentation parameters. \label{subfig:RotCIFARaug}}
	\end{subfigure}
	\begin{subfigure}[b]{0.495\linewidth}
	    \centering
	    \includegraphics[width=1.0\linewidth]{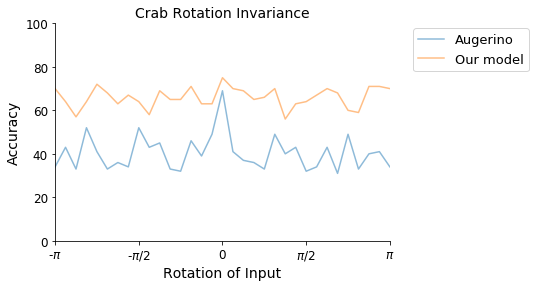}
	    \caption{Model invariance. \label{subfig:RotCIFARinv}}
	\end{subfigure}

	\captionsetup{width=\capwidth\linewidth}
	\caption{We train standard Augerino and our class conditional model on selected classes from the unmodified CIFAR-100 dataset. (\subref{subfig:RotCIFARaug}) Learned augmentation parameters over 50 training epochs. (\subref{subfig:RotCIFARinv}) Test accuracy vs.\ orientation on the crab images in the CIFAR-100 testing data.}\label{fig:RotCIFAR}
\end{figure}

Figure \ref{fig:RotCIFAR} demonstrates that our class conditional model does indeed learn larger augmentation parameters for classes that are suspected to have more rotation invariance. Two classes showed different results than we expected: the model learned little rotation invariance for clocks and more than expected for sea images. Upon further inspection, many of the CIFAR-100 images of clocks are clock towers at a standard orientation or have other features that are not rotationally invariant. On the other hand, some of the sea images could look similar when rotated by 180 degrees if the sea and sky look similar in that image. The advantage of our class conditional model is that it learns these invariances, even if we have no prior knowledge of them or they are different than expected.

We also show how each model performs on images of crabs, a class that appears at many orientations. We plot the models' accuracies on all crab images from the CIFAR-100 test set rotated at 32 evenly spaced orientations in the range $[-\pi,\pi]$. On this class, our model outperforms Augerino at every orientation. Augerino learns to rotate CIFAR-100 images at angles less than $0.16$ radians, or approximately in the range $[-\pi/20,\pi/20]$. It is likely that this small amount of augmentation limits Augerino's performance on classes with a lot of rotational symmetry. Moreover, our model's accuracy on crab images appears to vary less as orientation changes, suggesting that our model has more invariance to rotations of crabs. Augerino does attain 77.6\% accuracy on all 10 selected classes from CIFAR-100 while our model has 72.5\% accuracy. This accuracy gap could likely be closed by using more expressive network architectures and training the models longer. However, the standard Augerino model is not able to learn the class conditional invariant structure of the data like our model does.

\section{Limitations and Future Work} \label{sec:limitations}

Our model gives a method for learning class conditional invariances, but only provides one approach to this problem. Since we have shown how class conditional invariances can be present in real data and can improve model robustness, 
it is worth exploring other methods for implementing these types of invariances. One possibility draws from ideas of pooling outputs over different transformations of inputs or filters as in \cite{CW16a, CW16b, LSBP16, MVKT17, WC19, ZYQJ17} but using a weighted average over transformations in some data- or class-dependent way. We believe other methods for modeling class conditional invariance are a promising direction for future work.

Although our model is able to learn class conditional invariances from training data alone, it is still limited in scope. In some classification problems, invariance may depend on some features of the input data rather than class label. For example, color invariances may change for different breeds of dogs, but all dog breeds may be labeled as a single class. To address this challenge, we would have to design a model that conditions invariance on some other features of the input data. 



\section{Conclusions} \label{sec:conclusions}

In this paper, we study the problem of learning class conditional invariances from training data. To solve this problem, we extend the Augerino framework of learning invariances to allow for different data augmentation distributions for each class. We train our model on several datasets that motivate the problem of class conditional invariance and see that our model accurately learns different augmentations and generalizes well. Other approaches to learning class conditional invariances are worth exploring, and our ideas would be interesting to extend to other types of data-dependent invariances in order to enhance the automation and generalization of machine learning models.

\section*{Acknowledgements} \label{sec:acknowledgements}

This material is based upon work supported by the National Science Foundation Graduate Research Fellowship Program under Grant No.\ DGE-1650112. Any opinions, findings, and conclusions or recommendations expressed in this material are those of the author(s) and do not necessarily reflect the views of the National Science Foundation.

SM is funded by grant NSF DGE GRFP \#1650112.

This work used the Extreme Science and Engineering Discovery Environment (XSEDE) \cite{XSEDE} Expanse GPU through allocation TG-CIS200046. XSEDE is supported by National Science Foundation grant number ACI-1548562.

\end{document}